# Pathway Lasso: Estimate and Select Sparse Mediation Pathways with High Dimensional Mediators

Yi Zhao and Xi Luo

Department of Biostatistics, Brown University

**Abstract**

In many scientific studies, it becomes increasingly important to delineate the causal pathways through a large number of mediators, such as genetic and brain mediators. Structural equation modeling (SEM) is a popular technique to estimate the pathway effects, commonly expressed as products of coefficients. However, it becomes unstable to fit such models with high dimensional mediators, especially for a general setting where all the mediators are causally dependent but the exact causal relationships between them are unknown. This paper proposes a sparse mediation model using a regularized SEM approach, where sparsity here means that a small number of mediators have nonzero mediation effects between a treatment and an outcome. To address the model selection challenge, we innovate by introducing a new penalty called *Pathway Lasso*. This penalty function is a convex relaxation of the non-convex product function, and it enables a computationally tractable optimization criterion to estimate and select many pathway effects simultaneously. We develop a fast ADMM-type algorithm to compute the model parameters, and we show that the iterative updates can be expressed in closed form. On both simulated data and a real fMRI dataset, the proposed approach yields higher pathway selection accuracy and lower estimation bias than other competing methods.

# 1 Introduction

Causal mediation analysis is widely applied in social, economic and biological sciences to assess the effect of a treatment or exposure on an interested outcome passing through intermediate variables (mediators). It becomes increasingly popular to study the decomposition of the treatment effect on the outcome through multiple mediation pathways. This paper studies the problem of pathway selection and effect estimation under the setting of a large number of pathways, which can be larger than the sample size.

Several recent studies focus on the analysis of causal pathways with two mediators (Imai and Yamamoto, 2013; Daniel et al., 2014; VanderWeele et al., 2014; VanderWeele and Vansteelandt, 2014). For cases with more than two mediators, methods have been developed to estimate the decomposed mediation effects through bootstrap resampling (Preacher and Hayes, 2008), mediation formula (Wang et al., 2013), and estimating equations (Zhao et al., 2014). All these methods require fitting regression-type models with all the mediators as predictors, therefore, the estimates are unstable when the number of mediators is close or larger than the sample size. To circumvent this challenge, Huang and Pan (2015) proposed marginal mediation models for genetic data where the mediators after transformation are independent. In this study, we propose an alternative representation of the causal mechanisms with large number of causally dependent mediators under the structural equation modeling (SEM) framework, and introduce a novel convex penalty to select the mediation pathways and estimate causal mediation effects simultaneously. Our proposed model can be directly applied on the original data without transformation.

This paper is organized as follows. In Section 2, we introduce our pathway model, and discuss its causal assumptions and connection to an existing model. We introduce a novel $\ell_1$-type penalty on the pathway effects in Section 3. To estimate the parameters, an alternating direction method of multipliers (ADMM) combined with augmented Lagrangian algorithm is developed. In Section 4, we compare the performance of our approaches with the marginal SEM method and a two-stage lasso approach through simulation studies. The proposed methods are applied on an open-source functional magnetic resonance imaging (fMRI) dataset in Section 5. Section 6 summarizes the paper with some discussions.

## 2 Model

### 2.1 Multiple Mediator Models

Ideally, when the causal ordering of the mediators are known, a model for multiple mediators is plotted in Figure 1a (Daniel et al., 2014). In the figure, the mediators are causally related in such a

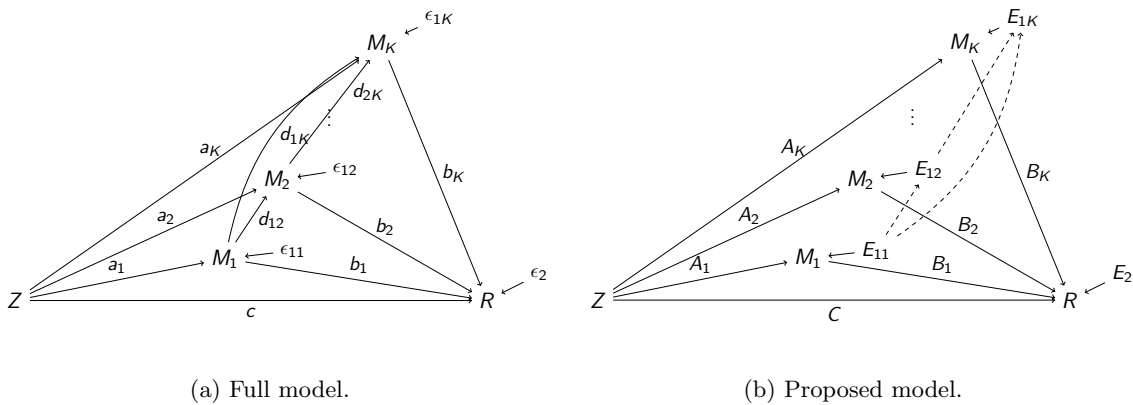

(a) Full model.  (b) Proposed model.

Figure 1: Causal diagram with multiple dependent mechanisms. $Z$ is the treatment variable, $R$ is the outcome variable, and $M_j$'s are the mediators.

way that "earlier" $M$'s (with smaller subscript) may affect "later" ones. The SEM representation

of this model is

$$M_1 = Za_1 + \epsilon_{11},$$

$$M_2 = Za_2 + M_1 d_{12} + \epsilon_{12},$$

$$\vdots \qquad (2.1)$$

$$M_K = Za_K + M_1 d_{1K} + M_2 d_{2K} + \cdots + M_{K-1} d_{K-1,K} + \epsilon_{1K},$$

$$R = Zc + M_1 b_1 + M_2 b_2 + \cdots + M_K b_K + \epsilon_2,$$

where $Z$ is a column vector of $n$ treatment assignment, $R$ and $M_1, \ldots, M_K$ are the corresponding outcome and mediator columns; $a_1, \ldots, a_K$, $b_1, \ldots, b_K$, $c$, $d_{12}, \ldots, d_{K-1,K}$ are the coefficients of interest; and $\epsilon_{11}, \ldots, \epsilon_{1K}$ and $\epsilon_2$ are the model error terms. We call this model as the full model.

There are two major issues with the full model (2.1). One is that the order of the mediators needs to be known, which requires strong prior knowledge or assumptions and is usually difficult to justify unless the mediators are observed in a temporal manner. In our application, the low temporal resolution of functional MRI provides limited information to determine the temporal ordering and thus the full model cannot be applied. More importantly, as shown by Daniel et al. (2014), the number of pathways and parameters grow exponentially as the number of mediators $K$ increases, and only small $K$ can be addressed in this full model, for example $K = 2$ in (Daniel et al., 2014).

We propose an alternative representation, which does not require the knowledge of the order of the mediators. When the order of the mediators is unknown, it is impossible to reveal the truly underlying causal relationships between the mediators. However, the "total" mediation effect of each mediator can be identified and is of interest for the purpose of pathway selection. We use a causal diagram in Figure 1b to represent this. In the figure, $E_{11}$ is the part of $M_1$ not explained by $Z$ and $E_{12}$ is the part of $M_2$ not explained by $Z$. Assume $M_1$ is in the earlier causal sequence and may have causal influence on $M_2$, then part of $E_{12}$ is explained by $E_{11}$ and therefore, they are not independent. Based on this causal diagram (Figure 1b), the SEM representation is

$$M_1 = ZA_1 + E_{11},$$
$$\vdots$$
$$M_K = ZA_K + E_{1K},$$
$$R = ZC + M_1B_1 + \cdots + M_KB_K + E_2,$$
(2.2)

where $A_1, \ldots, A_K$, $B_1, \ldots, B_K$ and $C$ are the coefficients of interest; and $E_{11}, \ldots, E_{1K}$ and $E_2$ are the model error terms.

## 2.2 Causal assumptions

We first discuss the causal interpretation of the full model (2.1). To assess it, we put model (2.1) under Rubin's potential outcome framework (Rubin, 2005). We impose the following assumptions:

**(A1)** stable unit treatment value assumption (SUTVA);

**(A2)** model (2.1) correctly specified;

**(A3)** the observed outcome is one realization of the potential outcome with observed treatment assignment $z$;

**(A4)** randomized treatment $Z$ with $0 < \mathbb{P}(Z = z) < 1$;

**(A5)** sequential ignorability with multiple causally dependent mediators.

These assumptions are standard regularity assumptions in causal mediation inference (Rubin, 1978; Holland, 1988; Imai et al., 2010; Imai and Yamamoto, 2013; VanderWeele, 2015). In this study, we assume that there is no interaction either between the treatment and the mediators or among the mediators.

## 2.3 Relationship between the Full Model and the Proposed Model

In this section, we discuss the relationship between the full model (2.1) (Figure 1a) and our proposed model (2.2) (Figure 1b). Comparing these two models, we can have the following relationships between the coefficients and the error terms,

$$A = A\Delta + a, \quad B = b, \quad C = c, \quad E_1 = E_1\Delta + \epsilon_1,$$

where $A = (A_1, \ldots, A_K)_{1 \times K}$, $a = (a_1, \ldots, a_K)_{1 \times K}$, $B = (B_1, \ldots, B_K)_{K \times 1}^\top$, $b = (b_1, \ldots, b_K)_{K \times 1}^\top$; $E_1 = (E_{11}, \ldots, E_{1K})_{n \times K}$, $\epsilon_1 = (\epsilon_{11}, \ldots, \epsilon_{1K})_{n \times K}$; and

$$\Delta = \begin{pmatrix} 0 & d_{12} & d_{13} & \cdots & d_{1K} \\ & 0 & d_{23} & \cdots & d_{2K} \\ & & \ddots & \ddots & \vdots \\ & & & \ddots & d_{K-1,K} \\ & & & & 0 \end{pmatrix}_{K \times K}$$

is the adjacency matrix of the mediators, which is an upper-triangular matrix. Since the diagonal of $(\mathbf{I}_K - \Delta)$ are all ones without any zero entries, it is invertible, where $\mathbf{I}_K$ is the $K$-dimensional identity matrix. We have the relationship between the coefficients and the error terms as $A = a(\mathbf{I}_K - \Delta)^{-1}$ and $E_1 = \epsilon_1 (\mathbf{I}_K - \Delta)^{-1}$, respectively. Matrix $(\mathbf{I}_K - \Delta)^{-1}$ is called the influence matrix (Shojaie and Michailidis, 2010), where the $(l, j)$ element represents the influence of mediator $M_l$ on mediator $M_j$ ($l < j$) with a self-influence of one when $l = j$. $(\mathbf{I}_K - \boldsymbol{\Delta})^{-1}$ is an upper-triangular matrix as well, and $A_j$ is the summation of the influence of first ($j-1$) mediators multiplied by the effect of $Z$ on $M_l$'s ($l < j$) and the direct effect of $Z$ on $M_j$. Therefore, $A_j$ can be interpreted as the total effect of treatment $Z$ on mediator $M_j$, regardless of the underlying causal relationship between the "earlier" $M_l$'s ($l < j$). As $B = \boldsymbol{b}$ and $C = c$, the interpretation of these coefficients in the proposed model is the same as in the full model. Thus, $A_j B_j$ is the total mediation effect of $M_j$ when it is the "last" mediator in the causal pathway from $Z$ to $R$. Imai and Yamamoto (2013) defined the mediation effect in a similar way under the situation of two mediators.

Under Assumption (A5), $\epsilon_{11}, \ldots, \epsilon_{1K}$ and $\epsilon_2$ are mutually independent, and thus $E_{1j}$ is independent of $E_2$ for each $j = 1, \ldots, K$. Assume $\text{Var}(\epsilon_{1j}) = \xi_{1j}^2 \mathbf{I}_n$, then

$$\text{Cov}\left[\text{vec}(\boldsymbol{\epsilon}_1)\right] = \text{diag}\left\{\xi_{11}^2, \ldots, \xi_{1K}^2\right\} \otimes \mathbf{I}_n \triangleq \boldsymbol{\Xi} \otimes \mathbf{I}_n,$$

$$\text{Cov}\left[\text{vec}(E_1)\right] = \left(\mathbf{I}_K - \boldsymbol{\Delta}^\top\right)^{-1} \boldsymbol{\Xi} \left(\mathbf{I}_K - \boldsymbol{\Delta}\right)^{-1} \otimes \mathbf{I}_n \triangleq \Sigma_1 \otimes \mathbf{I}_n,$$

where $\otimes$ is the Kronecker product operator and $\mathbf{I}_n$ is the $n \times n$ identity matrix. $\boldsymbol{\Xi}$ is a diagonal matrix and $(\mathbf{I}_K - \boldsymbol{\Delta})$ is an upper triangular matrix with all diagonal entities as one, the same as its inverse $(\mathbf{I}_K - \boldsymbol{\Delta})^{-1}$. $\left(\left(\mathbf{I}_K - \boldsymbol{\Delta}^\top\right)^{-1} \boldsymbol{\Xi} \left(\mathbf{I}_K - \boldsymbol{\Delta}\right)^{-1}\right)$ is the LDL decomposition of matrix $\Sigma_1$. When $\Sigma_1$ is symmetric positive-definite, this decomposition is unique. Therefore, we can obtain the causal relationship between the mediators by estimating $\Sigma_1$ and $A$ using model (2.2). When matrix $\boldsymbol{\Delta}$ is a zero matrix, meaning that given the treatment assignment $Z$, the "earlier" mediators have no impact on the "later" mediators (Imai and Yamamoto (2013) called them as causally independent mediators), $\Sigma_1 = \boldsymbol{\Xi}$ is a diagonal matrix.

For high-dimensional mediators ($K > n$), the sample covariance estimate for $\Sigma_1$ is not of full rank. We will specify a diagonal matrix model to estimate $\Sigma_1$. This modeling assumption approximately holds when $\boldsymbol{\Delta}$ is a sparse matrix and $\Sigma_1$ is thus a diagonally dominant matrix. In the simulation study in Section 4, the robustness of this assumption is examined.

## 3  Method

Let $M = (M_1, \ldots, M_K)_{n \times K}$, model (2.2) can be written into matrix form as

$$M = ZA + E_1,$$
$$R = ZC + MB + E_2. \tag{3.1}$$

Baron and Kenny (1986) considered a third equation $R = ZC' + E'$ with $C'$ as the total effect of $Z$ on $R$, which can be decomposed into indirect effect $(AB)$ and direct effect $(C)$, where $AB = \sum_{j=1}^{K} A_j B_j$ is the summation of the indirect effect of each $Z \to M_j \to R$ pathway. We consider the case that the outcome and mediators are continuous, and assume the model errors to be normally distributed,

$$\text{vec}(E_1) \sim \mathcal{N}_{n \times K}\left(\mathbf{0}, \Sigma_1 \otimes \mathbf{I}_n\right), \quad E_2 \sim \mathcal{N}_n\left(\mathbf{0}, \sigma_2^2 \mathbf{I}_n\right).$$

### 3.1  A *Pathway Lasso* method

We first define the loss function as

$$\ell = \text{tr}\left[W_1 (M - ZA)^\top (M - ZA)\right] + w_2 (R - ZC - MB)^\top (R - ZC - MB), \tag{3.2}$$

where $W_1 \succ 0$ (positive-definite) is the weight matrix for the mediator models, and $w_2 > 0$ is the weight for the outcome model. This loss function is convex in $(A, B, C)$.

To estimate and select the pathway effects $A_j B_j$, for $j = 1, \ldots, K$, we propose to minimize the following penalized criterion

$$f(A, B, C) = \frac{1}{2}\ell + \lambda \left[ \sum_{j=1}^{K} \left( |A_j B_j| + \phi(A_j^2 + B_j^2) \right) + |C| \right] + \omega \left[ \sum_{j=1}^{K} (|A_j| + |B_j|) \right] \quad (3.3)$$

$$= \frac{1}{2}\ell + \lambda P_1(A, B, C) + \omega P_2(A, B) \quad (3.4)$$

where $\lambda, \phi, \omega > 0$ are tuning parameters. The first penalty term $P_1$ aims to stabilize and shrink the estimates for the pathway effects $A_j B_j$, and the second term $P_2$ aims to provide additional shrinkage on the individual effects $A_j$ and $B_j$. The combination of two penalty terms is similar to the strategy in elastic net. It should also be noted that our method will also work if the tuning parameters vary with $j$. We here use fixed parameters for simplicity because our data are standardized.

A interesting case of the penalty $P_1$ is when $\phi = 0$, which intuitively shrinks the pathway effects $|A_j B_j|$ and $|C|$ towards zero via a Lasso-type penalty. However, $P_1$ is not convex under this setting. We prove that $P_1$ is convex if and only if $\phi \geq 1/2$.

**Theorem 3.1.** *For $a, b \in \mathbb{R}$, if and only if when $\phi \geq 1/2$,*

$$v(a, b) = |ab| + \phi(a^2 + b^2) \quad (3.5)$$

*is a convex function. When $\phi > 1/2$, it is strictly convex.*

Figure 2 shows the 3D plot of three different penalty functions and the contour plot of penalty functions with different choices of $\phi$. From the figures, we can see that $|ab|$ is not a convex function while $|ab| + (a^2 + b^2)/2$ is. The contour plot indicates that $\phi$ determines the convexity of the penalty function. The penalty $P_1$ is non-differentiable at the points where $ab = 0$. The contour of $P_1$ approaches the $\ell_2$ (or ridge) penalty when $\phi \to \infty$, and it approaches the $\ell_1$ penalty when $\phi = 1/2$. In Section 4, we will examine the choice of $\phi$ through simulation studies.

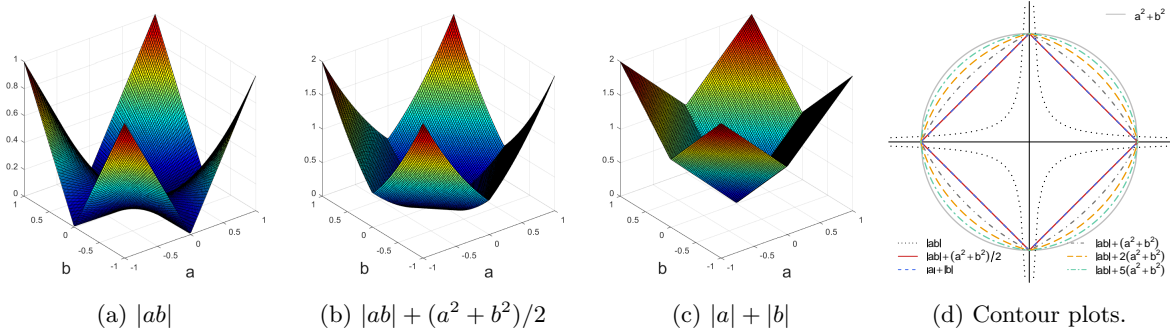

(a) $|ab|$  (b) $|ab| + (a^2 + b^2)/2$  (c) $|a| + |b|$  (d) Contour plots.

Figure 2: Penalty functions.

## 3.2  An alternating direction method of multipliers with an equality constraint

The objective function $f$ consists of two parts, *i)* the differentiable loss function $\ell/2$; and *ii)* the non-differentiable penalty function. We propose to employ the alternating direction method of multipliers (ADMM), which is well suited to large-scale statistical problem (Boyd et al., 2011). Let

$$X = \begin{pmatrix} Z & M \end{pmatrix}, \quad \Theta = \begin{pmatrix} 1 & A_1 & \cdots & A_K \end{pmatrix}, \quad D = \begin{pmatrix} C & B_1 & \cdots & B_K \end{pmatrix}^\top$$

$$\Phi = \mathrm{diag}\{0, \phi \mathbf{1}_K\}, \quad \Omega_1 = \mathrm{diag}\{0, W_1\}, \quad J = \begin{pmatrix} 0 & \mathbf{1}_K^\top \end{pmatrix}^\top,$$

where $\mathbf{1}_K$ is a vector of ones in $\mathbb{R}^K$. The ADMM form of problem (3.3) is

$$\begin{aligned}
\text{minimize} \quad & u(\Theta, D) + v(\alpha, \beta), \\
\text{subject to} \quad & \Theta = \alpha, \\
& D = \beta, \\
& \Theta e_1 = 1,
\end{aligned} \tag{3.6}$$

where

$$u(\Theta, D) = \frac{1}{2}\mathrm{tr}\left[\Omega_1 \left(X - Z\Theta\right)^\top \left(X - Z\Theta\right)\right] + \frac{1}{2} w_2 \left(R - XD\right)^\top \left(R - XD\right) \tag{3.7}$$

is the differentiable loss function, and

$$v(\alpha, \beta) = \lambda \left[ |\alpha||\beta| + \text{tr}\left(\Phi \alpha^\top \alpha\right) + \text{tr}\left(\Phi \beta \beta^\top\right)\right] + \omega \left[|\alpha|J + |\beta|^\top J\right] \tag{3.8}$$

is the non-differentiable regularization function; $|\alpha| = (|\alpha_1|, \ldots, |\alpha_{K+1}|)_{1 \times (K+1)}$, $|\beta| = (|\beta_1|, \ldots, |\beta_{K+1}|)^\top_{(K+1) \times 1}$; and $e_1$ is the standard basis in $\mathbb{R}^{K+1}$ with the first element as one.

Following the ADMM approach, the augmented Lagrange function $\mathcal{L}$ is introduced to enforce the constraints as

$$\mathcal{L}(\Theta, D, \alpha, \beta, \rho, \nu_r) = u(\Theta, D) + v(\alpha, \beta) + \sum_{r=1}^{3} \left(\nu_r h_r(\Theta, D, \alpha, \beta) + \rho h_r^2(\Theta, D, \alpha, \beta)\right), \tag{3.9}$$

where $h_1(\Theta, D, \alpha, \beta) = \Theta - \alpha$, $h_2(\Theta, D, \alpha, \beta) = D - \beta$, and $h_3(\Theta, D, \alpha, \beta) = \Theta e_1 - 1$. Our algorithm

---
**Algorithm 3.1** An algorithm of solving problem (3.6) using augmented Lagrangian method.

Given the results from the $s$th step, for the $(s+1)$th step,

$$\begin{aligned}
\Theta^{(s+1)} &= \underset{\Theta}{\text{argmin}} \ \mathcal{L}(\Theta, D^{(s)}, \alpha^{(s)}, \beta^{(s)}, \rho, \nu_r^{(s)}), \\
D^{(s+1)} &= \underset{D}{\text{argmin}} \ \mathcal{L}(\Theta^{(s+1)}, D, \alpha^{(s)}, \beta^{(s)}, \rho, \nu_r^{(s)}), \\
\begin{pmatrix} \alpha^{(s+1)} \\ \beta^{(s+1)} \end{pmatrix} &= \underset{\alpha, \beta}{\text{argmin}} \ \mathcal{L}(\Theta^{(s+1)}, D^{(s+1)}, \alpha, \beta, \rho, \nu_r^{(s)}), \\
\nu_r^{(s+1)} &= \nu_r^{(s)} + 2\rho h_r(\Theta^{(s+1)}, D^{(s+1)}, \alpha^{(s+1)}, \beta^{(s+1)}).
\end{aligned} \tag{3.10}$$

---

iteratively updates the parameters in (3.9), which is summarized in Algorithm 3.1. The solutions for the $\Theta$ and $D$ updates can be written in explicit forms (Appendix B). The subproblem for updating the $(\alpha, \beta)$ can be decomposed into $K+1$ optimization problems for each coordinate. Each optimization problem is of the same form and has explicit solutions as shown by the following lemma.

Table 1: Solution of optimization problem (3.11) when $\lambda \neq 0$.

| | Condition | Solution | |
|---|---|---|---|
| | | $a$ | $b$ |
| (1) | $\phi_2\mu_1 - \lambda\mu_2 > \omega(\phi_2 - \lambda)$ <br> $\phi_1\mu_2 - \lambda\mu_1 > \omega(\phi_1 - \lambda)$ | $x_1/(\phi_1\phi_2 - \lambda^2)$ | $y_1/(\phi_1\phi_2 - \lambda^2)$ |
| (2) | $\phi_2\mu_1 + \lambda\mu_2 > \omega(\phi_2 - \lambda)$ <br> $\phi_1\mu_2 + \lambda\mu_1 < -\omega(\phi_1 - \lambda)$ | $x_2/(\phi_1\phi_2 - \lambda^2)$ | $y_2/(\phi_1\phi_2 - \lambda^2)$ |
| (3) | $\phi_2\mu_1 + \lambda\mu_2 < -\omega(\phi_2 - \lambda)$ <br> $\phi_1\mu_2 + \lambda\mu_1 > \omega(\phi_1 - \lambda)$ | $x_3/(\phi_1\phi_2 - \lambda^2)$ | $y_3/(\phi_1\phi_2 - \lambda^2)$ |
| (4) | $\phi_2\mu_1 - \lambda\mu_2 < -\omega(\phi_2 - \lambda)$ <br> $\phi_1\mu_2 - \lambda\mu_1 < -\omega(\phi_1 - \lambda)$ | $x_4/(\phi_1\phi_2 - \lambda^2)$ | $y_4/(\phi_1\phi_2 - \lambda^2)$ |
| (5) | $\|\mu_1\| > \omega$ <br> $\phi_1\|\mu_2\| - \lambda\|\mu_1\| \leq \omega(\phi_1 - \lambda)$ | $(\|\mu_1\| - \omega)\mathrm{sgn}(\mu_1)/\phi_1$ | $0$ |
| (6) | $\|\mu_2\| > \omega$ <br> $\phi_2\|\mu_1\| - \lambda\|\mu_2\| \leq \omega(\phi_2 - \lambda)$ | $0$ | $(\|\mu_2\| - \omega)\mathrm{sgn}(\mu_2)/\phi_2$ |
| (7) | else | $0$ | $0$ |

**Lemma 3.2.** *For the optimization problem*

$$\underset{a,b\in\mathbb{R}}{minimize}\ v(a,b) = \lambda|ab| + \omega|a| + \omega|b| + \frac{1}{2}\phi_1 a^2 + \frac{1}{2}\phi_2 b^2 - \mu_1 a - \mu_2 b, \qquad (3.11)$$

*where $\phi_i > \lambda \geq 0$, $\omega \geq 0$ and $\mu_i \in \mathbb{R}$ ($i = 1, 2$),*

- *if $\lambda = 0$, the solution is*

$$a = \frac{1}{\phi_1}\mathcal{S}_\omega(\mu_1), \quad b = \frac{1}{\phi_2}\mathcal{S}_\omega(\mu_2), \qquad (3.12)$$

  *where $\mathcal{S}_\omega(\mu) = \max\{|\mu| - \omega, 0\}\mathrm{sgn}(\mu)$ is the soft-thresholding function;*

- *if $\lambda \neq 0$, it can be solved by using Table 1.*

*In the table, all conditions are mutually exclusive, and*

$$x_1 = \phi_2(\mu_1 - \omega) - \lambda(\mu_2 - \omega), \quad y_1 = \phi_1(\mu_2 - \omega) - \lambda(\mu_1 - \omega),$$

$$x_2 = \phi_2(\mu_1 - \omega) + \lambda(\mu_2 + \omega), \quad y_2 = \phi_1(\mu_2 + \omega) + \lambda(\mu_1 - \omega),$$

$$x_3 = \phi_2(\mu_1 + \omega) + \lambda(\mu_2 - \omega), \quad y_3 = \phi_1(\mu_2 - \omega) + \lambda(\mu_1 + \omega),$$

$$x_4 = \phi_2(\mu_1 + \omega) - \lambda(\mu_2 + \omega), \quad y_4 = \phi_1(\mu_2 + \omega) - \lambda(\mu_1 + \omega).$$

In the lemma, $\phi_1$ and $\phi_2$ are required to be greater than $\lambda$ to ensure the convexity of the objective function as shown in Theorem 3.1. The solutions show that the shrinkage effect towards $ab = 0$, including cases where only one of the $(a, b)$ parameters is zero (conditions 5-6). This is different from imposing the group lasso penalty $\sqrt{a^2 + b^2}$ (Yuan and Lin, 2006), where both of the parameters are either zero or non-zero.

Figure 3 shows the solution from Lemma 3.2 under different scenarios. From the figure, one can see that Table 1 yields the exact solution of the problem under various situations. This figure also shows that an alternative and simple approach, coordinate descent along the $a$ and $b$ axes, may fail to converge to the global minimum.

### 3.2.1 Penalty term $P_1$

The Lasso penalty ($P_2$) has been studied extensively in the literature. We here study analytically the behavior of the penalty $P_1$. Table 2 lists the solutions with the penalty $P_1$ along, which corresponds to setting $\omega = 0$ in Table 1.

**Proposition 3.3.** *When $\phi_i = \kappa_i \lambda + \theta_i$, where $\kappa_i \geq 1$ is a constant, $\theta_i > 0$ and $\theta_i/\lambda \to 0$ as $\lambda \to \infty$ ($i = 1, 2$), problem (3.11) with $\omega = 0$ is minimized at $a = b = 0$ when $\lambda \to \infty$.*

As shown in Appendix B, $\phi_1 = \phi_2 = 2\lambda\phi + 2\rho$ in Algorithm 3.1, where $\phi \geq 1/2$ by Theorem 3.1

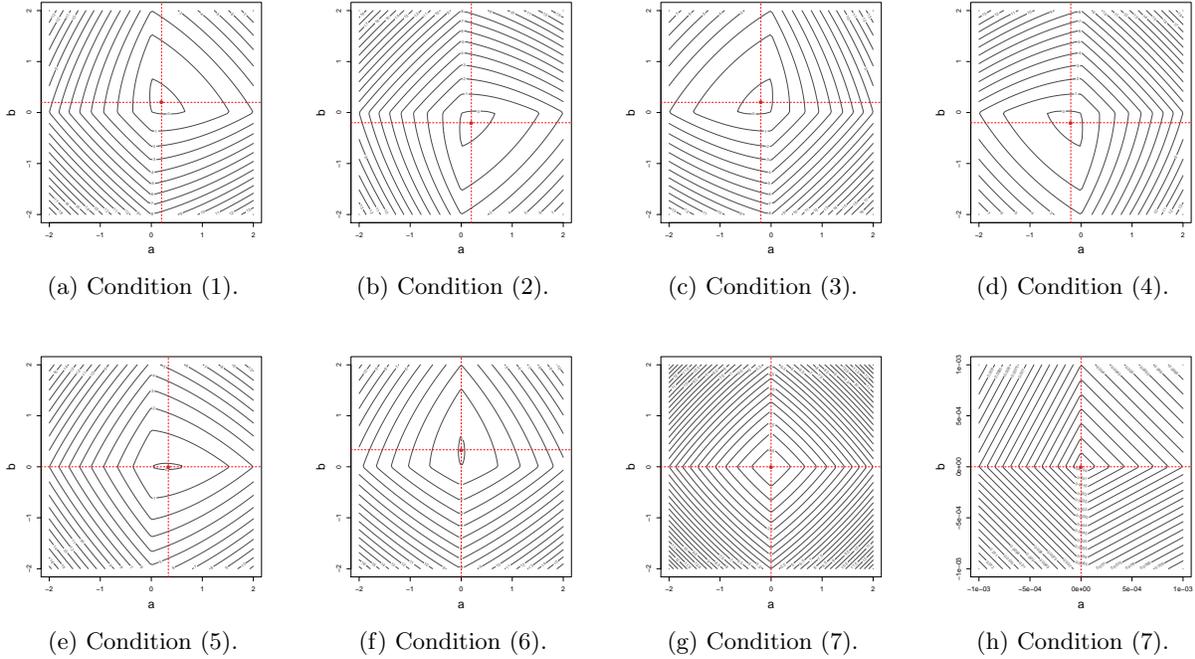

Figure 3: An example of solving problem (3.11) using Lemma 3.2 under eight scenarios corresponding to the seven conditions in Table 1. For all the cases, $\lambda = 1$, $\phi_1 = \phi_2 = 1.5$, and (a) $\omega = 1$, $\mu_1 = \mu_2 = 1.5$; (b) $\omega = 1$, $\mu_1 = 1.5$, $\mu_2 = -1.5$; (c) $\omega = 1$, $\mu_1 = -1.5$, $\mu_2 = 1.5$; (d) $\omega = 1$, $\mu_1 = \mu_2 = -1.5$; (e) $\omega = 1$, $\mu_1 = 1.5$, $\mu_2 = 0.15$; (f) $\omega = 1$, $\mu_1 = 0.15$, $\mu_2 = 1.5$; (g) $\omega = 1$, $\mu_1 = \mu_2 = 0$; (h) $\omega = 5$, $\mu_1 = \mu_2 = 1.5$. The red solid dot is the solution from the lemma.

Table 2: Solution of optimization problem (3.11) when $\omega = 0$.

| | Condition | Solution | |
|---|---|---|---|
| | | $a$ | $b$ |
| (1) | $(\phi_2\mu_1 - \lambda\mu_2)(\phi_1\mu_2 - \lambda\mu_1) > 0$ | $(\phi_2\mu_1 - \lambda\mu_2)/(\phi_1\phi_2 - \lambda^2)$ | $(\phi_1\mu_2 - \lambda\mu_1)/(\phi_1\phi_2 - \lambda^2)$ |
| (2) | $(\phi_2\mu_1 + \lambda\mu_2)(\phi_1\mu_2 + \lambda\mu_1) < 0$ | $(\phi_2\mu_1 + \lambda\mu_2)/(\phi_1\phi_2 - \lambda^2)$ | $(\phi_1\mu_2 + \lambda\mu_1)/(\phi_1\phi_2 - \lambda^2)$ |
| (3) | $|\mu_1| > 0$ and $\phi_1|\mu_2| - \lambda|\mu_1| \leq 0$ | $\mu_1/\phi_1$ | 0 |
| (4) | $|\mu_2| > 0$ and $\phi_2|\mu_1| - \lambda|\mu_2| \leq 0$ | 0 | $\mu_2/\phi_2$ |
| (5) | $\mu_1 = \mu_2 = 0$ | 0 | 0 |

and $\rho$ is the Lagrangian multiplier in our algorithm. In ADMM algorithms, $\rho$ can be fixed and increasing. This proposition shows that the solutions by our algorithm will go to zero when $\lambda \to \infty$, as long as $\rho/\lambda \to 0$. We will use fixed $\rho = 1$ for simplicity.

# 4 Simulation Study

In this section, we compare our Pathway Lasso (PathLasso) method with a marginal SEM approach and a two-stage lasso (TSLasso) method through simulation studies.

In the marginal SEM approach, Baron-Kenny (BK) (Baron and Kenny, 1986) mediation analysis is applied to each mediator separately. When the mediators are orthogonal or causally independent (VanderWeele, 2015), the parameters are equivalent to our multiple mediator model (3.1). The BK estimators are biased (Zhao and Luo, 2014) under the setting without causal independence. The pathway effects or the product estimators are tested by the delta method (Sobel, 1982) and significant pathways are selected based on false discovery rates (Benjamini and Hochberg, 1995) smaller than a nominal value of 5%. The two-stage lasso method uses $\ell_1$ penalized regression for each equation in (3.1), which is equivalent to setting $\lambda = 0$ and $\omega > 0$ in our method.

In the simulation study, we generate $n = 50$ samples and vary the number of mediator with $K = 50, 200$. For our proposed method, we consider three approaches with $a)$ $\omega = 0$; $b)$ $\omega = 0.1\lambda$; and $c)$ $\omega = \lambda$. Tuning parameter $\lambda$ changes from $10^{-6}$ to $10^2$. We let $\phi$ vary from $\{0.5, 1, 2, 5\}$. The data is standardized first. We set $W_1$ as an identity matrix and $w_2$ equal to one. To examine the robustness of this choice, we set $\Sigma_1$ to be a sparse matrix with sparsity $(1 - 1/K)$ (the proportion of nonzero off-diagonal entries is $1/K$) and vary the correlation value $\rho_M$ from $\{0, \pm 0.4\}$. We use $10^{-3}$ as a cut-off to decide if a causal pathway is selected or not. To compare the performance of various methods without setting the tuning parameters or p-value thresholds, we first employ the following metrics: (1) receiver operating characteristic (ROC) curves and $F_1$-score, and (2) the mean squared error (MSE) of $AB$ estimates. For a fair comparison, we compare the $F_1$ scores and MSE under the same number of selected mediators and the same $\ell_1$ norm of the estimated pathway effects for all methods, respectively.

Figure 4 compares the performance under different $\phi$ values when $\omega = 0$ and $\rho_M = 0$. From the figure, we see that our method is not sensitive to the choice of $\phi$ in selecting causal mediation pathways. When $K = 200$, the area under the curve (AUC) of $\phi = 5$ is significantly lower than the other choices. When $\phi = 2$, the method yields lower MSE in estimating $AB$, and thus we fix $\phi = 2$ for the following simulations.

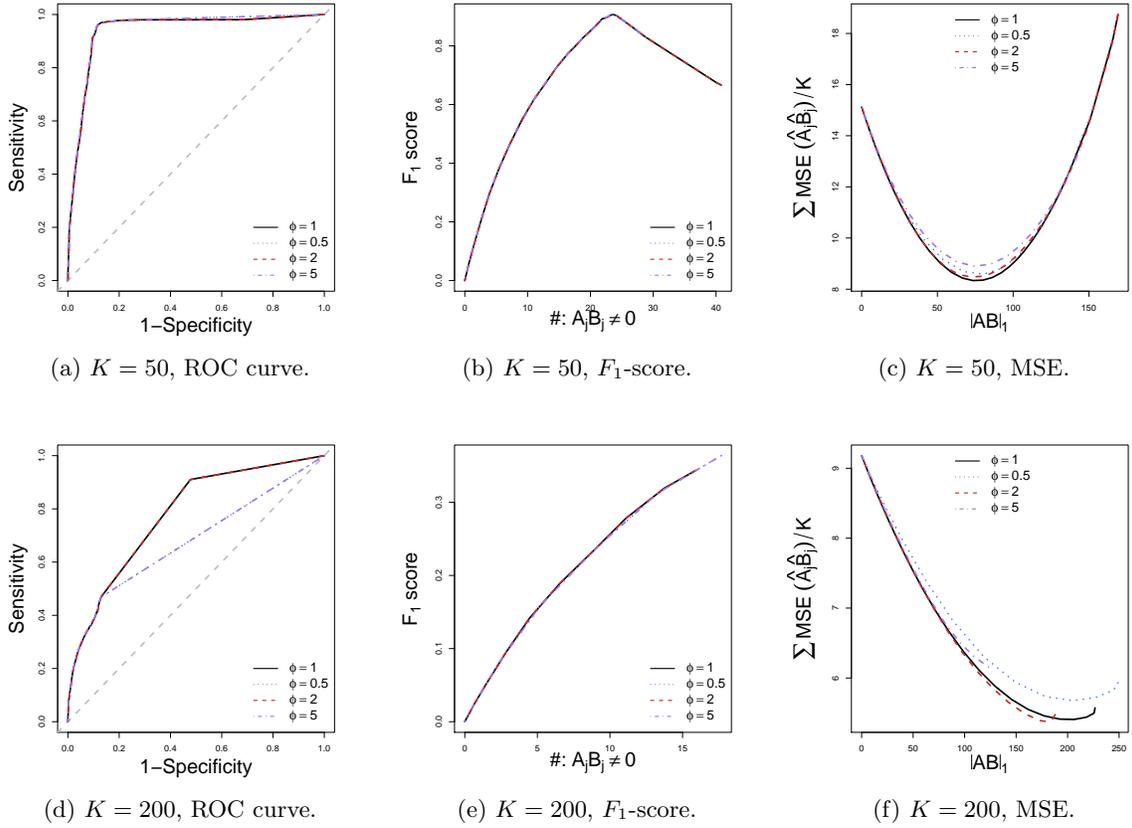

(a) $K = 50$, ROC curve.  (b) $K = 50$, $F_1$-score.  (c) $K = 50$, MSE.

(d) $K = 200$, ROC curve.  (e) $K = 200$, $F_1$-score.  (f) $K = 200$, MSE.

Figure 4: Performance comparison of PathLasso with $\omega = 0$ under different $\phi_j$ values when $\rho_M = 0$. (Black solid line: $\phi_j = 1$, blue dotted line: $\phi_j = 1/2$, red dashed line: $\phi_j = 2$, purple dotdash line: $\phi_j = 5$.)

Figure 5 compares the performance of all the considered methods with/without correlation between the mediators. The ROC curves of the BK method are almost the same as the diagonal line, indicating that even when the mediators are causally independent, this multiple testing for marginal mediation effect approach loses the power of identifying the significant causal mechanisms. When $K = 50$, the performance of TSLasso approach and the PathLasso approaches in selecting causal pathways are both good with TSLasso achieving slightly higher in terms of AUC and $F_1$-score. When the number of mediator increases to 200, both PathLasso with $\omega = 0$ and $\omega = 0.1\lambda$ outperform TSLasso approach with higher AUC. All three PathLasso approaches attain higher $F_1$-score than TSLasso approach does. When comparing the MSE in estimating the mediation effects $A_j B_j$'s, PathLasso with $\omega = 0$ achieves the lowest MSE followed by PathLasso with $\omega = 0.1\lambda$, PathLasso with $\omega = \lambda$ and TSLasso for both $K = 50$ and $K = 200$. This indicates that our proposed method of introducing $\ell_1$ penalty on the product term can significantly reduces the estimation bias in the mediation effects. For all three comparison metrics, the superior performance of PathLasso with $\omega = 0$ over TSLasso becomes more significant for larger $K$. The figure also demonstrates that our method is robust to the situation when the errors of the mediator models are correlated.

## 4.1 Choice of $\lambda$

This section compares the performance when the tuning parameters are chosen by 10-fold cross validation. Because the performance under different correlation $\rho_M$ is similar in the previous section, we only present the results when $\rho_M = 0$. Table 3 lists the average $F_1$ scores and MSEs from different methods. For both $K$s, PathLasso with $\omega = 0$ approach achieves the highest $F_1$-score and the lowest MSE, and all three PathLasso approaches perform better than TSLasso approach does. When $K = 200$, the improvement of PathLasso with $\omega = 0$ approach in $F_1$-score is about 200% (with a value of 0.583) compared to the two-stage lasso approach (whose $F_1$-score is 0.197).

## 5 An fMRI Study

We apply our method on a task-fMRI dataset available from the Human Connectome Project (HCP)[1]. We use the task fMRI data from a healthy female subject aged between 26 to 30, and the task is the language processing task. The language task was studied before, see the details in Binder et al. (2011). The HCP dataset consists of two runs (LR and RL phase encoding), and each run consists of randomized interleaves of four blocks of story task and four blocks of math task. The subject will hear brief auditory stories under the story task, and arithmetic questions under the math task. In this study, we define a 5mm radius spherical ROI in the Brodmann area 44 (MNI coordinate: -49, 9, 5) suggested by Heim et al. (2009) as the outcome region ($R$), which has been well studied in the literature. The goal is to study the brain pathways from the stimuli (story/math presentation) to the outcome region activity.

---

[1]HCP: http://www.humanconnectomeproject.org/

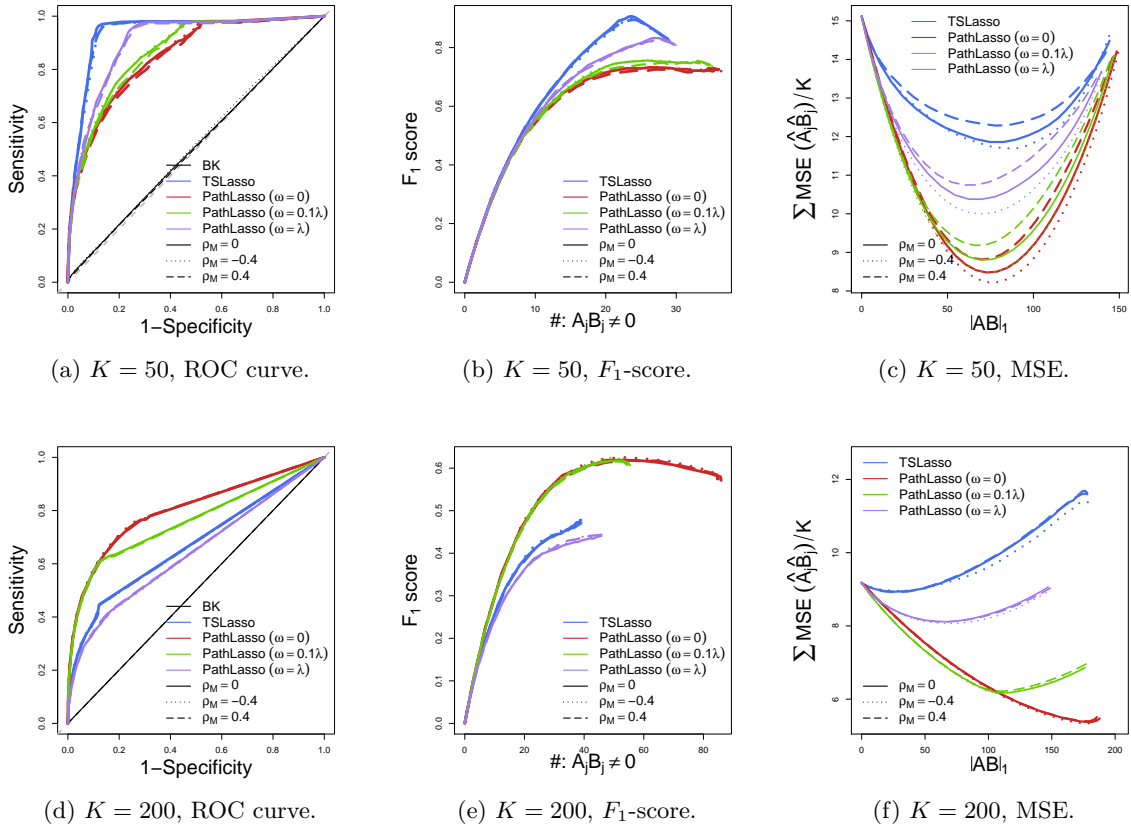

Figure 5: Performance comparison under each $\rho_M$. For PathLasso methods $\phi_j = 2$. (Black line: BK method, blue line: TSLasso method, red line: PathLasso with $\omega = 0$, green line: PathLasso with $\omega = 0.1\lambda$, purple line: PathLasso with $\omega = \lambda$. Solid line: $\rho_M = 0$, dotted line: $\rho_M = -0.4$, dashed line: $\rho_M = 0.4$.)

Table 3: Performance comparison with chosen tuning parameter by 10-fold cross-validation when the mediators are causally independent. For PathLasso approaches, $\phi_j = 2$. (Values in parentheses are standard errors.)

|  | Method | $K = 50$ | $K = 200$ |
| --- | --- | --- | --- |
| $F_1$-score | TSLasso | 0.658 (0.225) | 0.166 (0.087) |
|  | PathLasso ($\omega = 0$) | **0.725** (0.048) | **0.589** (0.061) |
|  | PathLasso ($\omega = 0.1\lambda$) | 0.748 (0.055) | 0.571 (0.078) |
|  | PathLasso ($\omega = \lambda$) | 0.681 (0.149) | 0.170 (0.080) |
| MSE | TSLasso | 13.105 (3.318) | 8.993 (0.771) |
|  | PathLasso ($\omega = 0$) | **9.381** (2.174) | **6.414** (0.622) |
|  | PathLasso ($\omega = 0.1\lambda$) | 9.709 (2.053) | 6.586 (0.547) |
|  | PathLasso ($\omega = \lambda$) | 11.775 (2.173) | 8.655 (0.265) |

We use the minimally processed data provided by the HCP. Due to the smoothness at the voxel level, we apply group independent component analysis (ICA) to both runs, using FSL MELODIC (Jenkinson et al., 2012; Woolrich et al., 2009; Smith et al., 2004). The analysis yields 76 independent components as mediators. We compare the performance of TSLasso and PathLasso because BK has the worst performance in our simulation studies. Again, we choose $\phi = 2$ in our method because its superior performance in our simulation studies.

To compare the methods without the truth, we compare the difference between the results from two runs. We use the $\ell_2$ difference to assess the pathway estimation accuracy, and the Jaccard index to assess the pathway selection accuracy. Figure 6 shows that PathLasso with $\omega = 0$ and $\omega = 0.1\lambda$ yield lower $\ell_2$ difference and higher Jaccard index values, for all tuning parameters or varying $\ell_1$-norm of the estimates, This indicates that our proposed method obtains more stable results in both the mediation effect estimates and the selection of nonzero mediation pathways. Using the tuning parameters selected by 10-fold cross validation for both methods, Table 4 shows that PathLasso with $\omega = 0$ has about 70% higher in the Jaccard index value than TSLasso, while the $\ell_2$ difference of PathLasso is slightly worse.

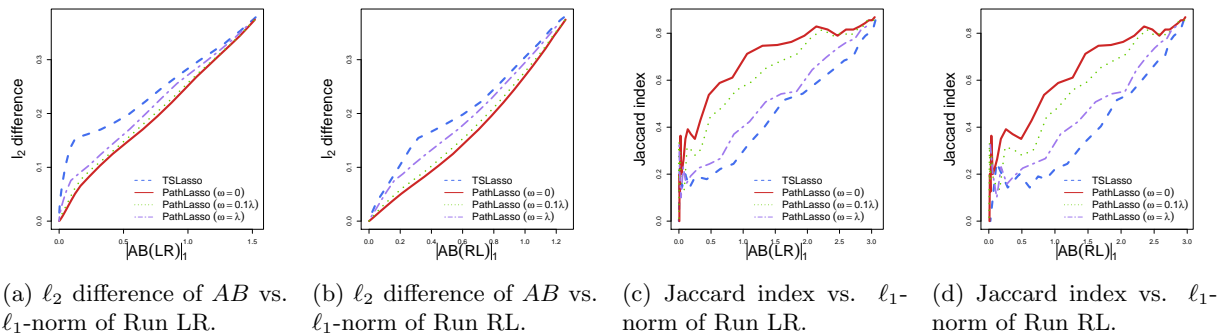

(a) $\ell_2$ difference of $AB$ vs. $\ell_1$-norm of Run LR.

(b) $\ell_2$ difference of $AB$ vs. $\ell_1$-norm of Run RL.

(c) Jaccard index vs. $\ell_1$-norm of Run LR.

(d) Jaccard index vs. $\ell_1$-norm of Run RL.

Figure 6: Performance comparison of the fMRI dataset. (Blue dashed line: TSLasso method, red solid line: PathLasso with $\omega = 0$, green dotted line: PathLasso with $\omega = 0.1\lambda$, purple dotdash line: PathLasso with $\omega = \lambda$.)

Table 4: Performance comparison with chosen tuning parameter by 10-fold cross-validation of the fMRI dataset.

|  | TSLasso | PathLasso | | |
| --- | --- | --- | --- | --- |
|  |  | $\omega = 0$ | $\omega = 0.1\lambda$ | $\omega = \lambda$ |
| $\ell_2$ difference | 0.260 | 0.306 | 0.298 | **0.258** |
| Jaccard index | 0.459 | **0.724** | 0.645 | 0.515 |

To ameliorate the bias, we refit our model using selected pathways without the penalty terms. Following Bunea et al. (2011), we use 500 bootstrapped samples to assess the significance of the pathways . IC 27 is found significant, and its pathway effects are consistently negative in both runs: $-0.063$ (95% CI: $(-0.107, -0.024)$, 31.7% mediated) in the LR run and $-0.038$ (CI: $(-0.075, -0.006)$, 13.4% mediated) in the RL run.

Figure 7(a) shows the IC maps of IC 27. On this map, the largest cluster with positive weights comes from the superior frontal gyrus (peak MNI coordinate: 2, 34, 60). This area was implicated in arithmetic processing before (Kesler et al., 2006) but the pathway effects has not been studied before in the whole brain. To visualize the pathway, we use this area as the representing region for the mediator map because its voxel weights are much larger than other areas, and the brain pathway is plotted in Figure 7(b).

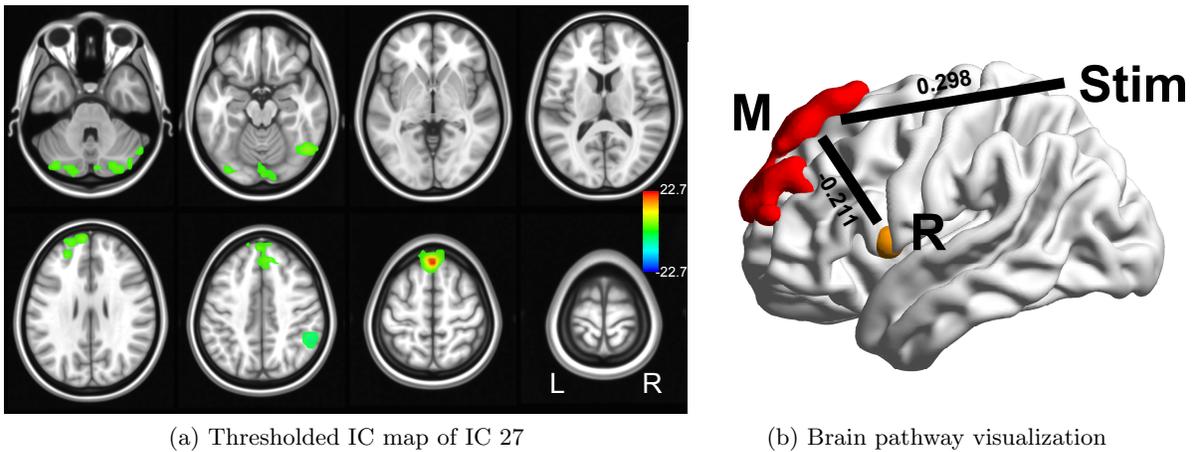

(a) Thresholded IC map of IC 27  (b) Brain pathway visualization

Figure 7: Brain maps of IC 27 weights, thresholded by equal false positive and negative probability.

## 6  Discussion

In this study, we propose a representation of "total mediation effect" for causally dependent mediators under SEM framework. We propose a novel convex penalty for shrinkage estimation and pathway selection. We develop an ADMM algorithm to estimate the parameters, and we provide the explicit solution to the iterative updates. The simulation studies indicate that our Pathway

Lasso method is robust and performs better than the marginal mediation approach in identifying significant causal mechanisms, compared with alternative methods. The numeric merits are further illustrated using a task-fMRI dataset, on which our Pathway Lasso shows higher replicability in both estimation and pathway selection.

# A  Proof of Theorem 3.1

*Proof.* To show the covexity of (3.5), we can write it in an equivalent form as

$$v(a,b) = |ab| + \phi(a^2 + b^2) = \max\left\{\frac{1}{2}(a+b)^2, \frac{1}{2}(a-b)^2\right\} + (\phi - \frac{1}{2})(a^2 + b^2). \tag{A.1}$$

Therefore, when $\phi \geq 1/2$, $v(a,b)$ is a convex function.

Let $(a_1, b_1) = (0, 1)$ and $(a_2, b_2) = (1, 0)$, for $\forall\, t \in [0, 1]$, we have

$$v(a_1, b_1) = v(a_2, b_2) = \phi, \quad \Rightarrow \quad tv(a_1, b_1) + (1-t)v(a_2, b_2) = \phi$$

$$a' = ta_1 + (1-t)a_2 = 1-t, b' = tb_1 + (1-t)b_2 = t, \quad \Rightarrow \quad v(a', b') = t(1-t) + \phi(2t^2 - 2t + 1)$$

$$\Rightarrow \quad v(a', b') - [tv(a_1, b_1) + (1-t)v(a_2, b_2)] = t(1-t)(1 - 2\phi)$$

∴ When $\phi < 1/2$, there exits $(a_1, b_1) = (0, 1)$, $(a_2, b_2) = (1, 0)$ such that for $\forall\, t \in (0, 1)$,

$$v\left(ta_1 + (1-t)a_2, tb_1 + (1-t)b_2\right) > tv(a_1, b_1) + (1-t)v(a_2, b_2).$$

Therefore, $\phi \geq 1/2$ is a necessary and sufficient condition of $v(a,b)$ being a convex function. □

# B  The iterative updates of Algorithm 3.1

*Proof.*

$$\mathcal{L}(\Theta, D, \alpha, \beta, \rho, \nu_r) = u(\Theta, D) + v(\alpha, \beta) + \sum_{r=1}^{3}\left(\nu_r h_r(\Theta, D, \alpha, \beta) + \rho h_r^2(\Theta, D, \alpha, \beta)\right)$$

Since $u(\Theta, D)$ is a smooth function of $(\Theta, D)$, we can find the updates by taking partial derivatives, and we have for the $(s+1)$th step,

$$\Theta^{(s+1)} = \left[Z^\top X\Omega_1 - \nu_1^{(s)\top} + 2\rho\alpha^{(s)} + (2\rho - \nu_3^{(s)})e_1^\top\right]\left[Z^\top Z\Omega_1 + 2\rho(\mathbf{I}_{K+1} + e_1 e_1^\top)\right]^{-1},$$

$$D^{(s+1)} = \left[w_2 X^\top X + 2\rho\mathbf{I}_{K+1}\right]^{-1}\left[w_2 X^\top R - \nu_2^{(s)} + 2\rho\beta^{(s)}\right]$$

$\Phi$ is a diagonal matrix, therefore, for the part related to $\alpha$ and $\beta$ (with given $\Theta$, $D$ and $\nu_r$'s), we have

$$v(\alpha, \beta) + \sum_{r=1}^{2}\left(\nu_r h_r(\Theta, D, \alpha, \beta) + \rho h_r^2(\Theta, D, \alpha, \beta)\right)$$

$$= \lambda\left[|\alpha||\beta| + \mathrm{tr}(\Phi\alpha^\top\alpha) + \mathrm{tr}(\Phi\beta\beta^\top)\right] + \omega\left(|\alpha|J + J^\top|\beta|\right) + (\Theta - \alpha)\nu_1 + \rho\parallel\Theta - \alpha\parallel_2^2 + \nu_2^\top(D - \beta) + \rho\parallel D - \beta\parallel_2^2$$

$$= \sum_{j=1}^{K+1}\left\{\lambda|\alpha_j\beta_j| + \omega J_j|\alpha_j| + \omega J_j|\beta_j| + \frac{1}{2}(2\lambda\Phi_j + 2\rho)\alpha_j^2 + \frac{1}{2}(2\lambda\Phi_j + 2\rho)\beta_j^2 - (2\rho\Theta_j + \nu_{1j})\alpha_j - (2\rho D_j + \nu_{2j})\beta_j + \mathrm{const}_j\right\},$$

where $\mathrm{const}_j = \rho\Theta_j^2 + \nu_{1j}\Theta_j + \rho D_j^2 + \nu_{2j}D_j$. To minimize this function over $\alpha$ and $\beta$ is equivalent to minimize it element-wise, and for each $j$, this can be achieved by using Lemma 3.2. $\square$

## C Proof of Lemma 3.2

*Proof.* When $\lambda = 0$, the problem is simplified into optimizing $\ell_1$ norm problem, and we can solve for the minimizers by soft-thresholding.

When $\lambda \neq 0$, conditions (1) to (4) are the cases when (1) $a > 0$ and $b > 0$, (2) $a > 0$ and $b < 0$, (3) $a < 0$ and $b > 0$, (4) $a < 0$ and $b < 0$, respectively, and the solutions are found by setting the gradient of the function (3.11) to zero under each case. Here we only show the proof of condition (5), and condition (6) can be proved analogously. When $b = 0$, the function is non-differentiable but we can find the subgradient $\nabla v(a, b)$ which is required to be in an interval, by the subgradient of the max of two functions in (A.1). In the same time, $v(a, 0)$ is minimized at $a = (|\mu_1| - \omega)\mathrm{sgn}(\mu_1)/\phi_1$ by soft-thresholding, and this is the solution if the interval requirement is satisfied, which translates to the condition $\phi_1|\mu_2| - \lambda|\mu_1| \leq \omega(\phi_1 - \lambda)$.

When $\omega = 0$, we obtain Table 2 using the same approach, and one can prove that the complimentary of the union of conditions (1)-(4) is $\mu_1 = \mu_2 = 0$ by discussing the sign of $\mu_1\mu_2$. $\square$

# D  Proof of Proposition 3.3

*Proof.* Suppose condition (1) is satisfied, i.e., $(\phi_2\mu_1 - \lambda\mu_2)(\phi_1\mu_2 - \lambda\mu_1) > 0$, if

(i)
$$\begin{cases} \phi_2\mu_1 - \lambda\mu_2 > 0 \\ \phi_1\mu_2 - \lambda\mu_1 > 0 \end{cases} \Rightarrow \frac{\lambda}{\phi_2}\mu_2 < \mu_1 < \frac{\phi_1}{\lambda}\mu_2 \Leftrightarrow \frac{\lambda}{\kappa_2\lambda + \theta_2}\mu_2 < \mu_1 < \frac{\kappa_1\lambda + \theta_1}{\lambda}\mu_2.$$

$$a = \frac{\phi_2\mu_1 - \lambda\mu_2}{\phi_1\phi_2 - \lambda^2} = \frac{(\kappa_2\mu_1 - \mu_2)\lambda + \theta_2\mu_1}{(\kappa_1\kappa_2 - 1)\lambda^2 + (\kappa_1\theta_2 + \kappa_2\theta_1)\lambda + \theta_1\theta_2}$$

$$b = \frac{\phi_1\mu_2 - \lambda\mu_1}{\phi_1\phi_2 - \lambda^2} = \frac{(\kappa_1\mu_2 - \mu_1)\lambda + \theta_1\mu_2}{(\kappa_1\kappa_2 - 1)\lambda^2 + (\kappa_1\theta_2 + \kappa_2\theta_1)\lambda + \theta_1\theta_2}$$

If $\kappa_1\kappa_2 = 1$, i.e., $\kappa_1 = \kappa_2 = 1$, when $\lambda \to \infty$, the inequality condition shows that $\mu_1 \to \mu_2$, and

$$a = \frac{(\mu_1 - \mu_2)\lambda + \theta_2\mu_1}{(\theta_1 + \theta_2)\lambda + \theta_1\theta_2} \xrightarrow{\lambda \to \infty} \frac{\mu_1 - \mu_2}{\theta_1 + \theta_2} = 0, \quad b = \frac{(\mu_2 - \mu_1)\lambda + \theta_1\mu_2}{(\theta_1 + \theta_2)\lambda + \theta_1\theta_2} \xrightarrow{\lambda \to \infty} \frac{\mu_2 - \mu_1}{\theta_1 + \theta_2} = 0$$

If $\kappa_1\kappa_2 > 1$, as $\lambda \to \infty$, $\mu_2/\kappa_2 < \mu_1 < \kappa_1\mu_2$, and we have

$$a \xrightarrow{\lambda \to \infty} 0, \quad b \xrightarrow{\lambda \to \infty} 0.$$

Analogously, we have the same conclusion for condition (1) when $\phi_2\mu_1 - \lambda\mu_2 < 0$ and $\phi_1\mu_2 - \lambda\mu_1 < 0$, as well as for condition (2). For conditions (3) and (4), it is obvious that when $\lambda \to \infty$, $a \to 0$ and $b \to 0$, respectively. □